\documentclass[letterpaper, 10 pt, conference]{ieeeconf}  

\IEEEoverridecommandlockouts                              

\usepackage{graphicx} 
\usepackage{amsmath} 
\usepackage{amssymb}  
\usepackage{cite}
\usepackage{textcomp}
\usepackage{tabularx}
\usepackage{gensymb}
\usepackage{breqn}
\usepackage{xcolor}
\usepackage{placeins}
\usepackage[]{multirow}
\usepackage{subcaption}
\usepackage{tikz}
\usepackage{geometry}
\usepackage{algpseudocode}
\usepackage{algorithm}
\usepackage{booktabs}
\algdef{SE}[DOWHILE]{Do}{doWhile}{\algorithmicdo}[1]{\algorithmicwhile\ #1}%

\usepackage[normalem]{ulem}

\title{\LARGE \bf
	Testing Robot System Safety by creating Hazardous Human Worker Behavior in Simulation
}

\newcommand{\breakingcomma}{%
	\begingroup\lccode`~=`,
	\lowercase{\endgroup\expandafter\def\expandafter~\expandafter{~\penalty0 }}}

\newcommand*\titleheader[1]{\gdef\@titleheader{#1}}

\newcommand\copyrighttext{%
	\footnotesize © 2021 IEEE.  Personal use of this material is permitted.  Permission from IEEE must be obtained for all other uses, in any current or future media, including reprinting/republishing this material for advertising or promotional purposes, creating new collective works, for resale or redistribution to servers or lists, or reuse of any copyrighted component of this work in other works.}
\newcommand\copyrightnotice{%
	\begin{tikzpicture}[remember picture,overlay]
		\node[anchor=north,yshift=-20pt] at (current page.north) {\fbox{\parbox{\dimexpr\textwidth-\fboxsep-\fboxrule\relax}{\copyrighttext}}};
	\end{tikzpicture}%
}

\author{Tom P. Huck, Christoph Ledermann, and Torsten Kr\"oger
	\thanks{\noindent This work was funded by the German Federal Ministry of Economics in the research project 'FabOS'.}
	\thanks{The authors are with the Intelligent Process Automation and Robotics Lab, Institute of Anthropomatics and Robotics (IAR-IPR), Karlsruhe Institute of Technology (KIT), 76131 Karlsruhe, Germany.\newline Corresponding author: Tom P. Huck
		({\tt\small tom.huck@kit.edu})}%
	\thanks{This is a preprint of a paper accepted for publication in the IEEE Robotics and Automation Letters (RA-L). The final published version may differ.}
}

\begin{document}

\maketitle
\thispagestyle{empty}
\pagestyle{empty}
\copyrightnotice

\begin{abstract}
We introduce a novel simulation-based approach to identify hazards that result from unexpected worker behavior in human-robot collaboration. Simulation-based safety testing must take into account the fact that human behavior is variable and that human error can occur. When only the \textit{expected} worker behavior is simulated, critical hazards can remain undiscovered. On the other hand, simulating \textit{all} possible worker behaviors is computationally infeasible. This raises the problem of how to find \textit{interesting} (i.e., potentially hazardous) worker behaviors given a limited number of simulation runs.
We frame this as a search problem in the space of possible worker behaviors. Because this search space can get quite complex, we introduce the following measures:
(1) Search space restriction based on workflow-constraints, 
(2) prioritization of behaviors based on how far they deviate from the nominal behavior, and
(3) the use of a risk metric to guide the search towards high-risk behaviors which are more likely to expose hazards. We demonstrate the approach in a collaborative workflow scenario that involves a human worker, a robot arm, and a mobile robot.
\end{abstract}

\section{Introduction}
Hazard analysis is a major challenge in human-robot collaboration (HRC). As HRC is increasingly deployed in industrial practice, effective methods are needed to ensure safety of human workers. While traditional robot systems maintain safety through spatial separation, HRC requires more sophisticated measures, like adaptive speed- and workspace limitations, area monitoring (e.g. by laser scanners and light curtains), collision detection, and more \cite{STD_ISOTS15066}. These measures must be configured appropriately with respect to workflow and robot properties like stopping time and collision potential. Furthermore, the measures must ensure safety not only for expected worker behaviors, but also for unforeseen or erroneous behaviors. To ensure that these requirements are met, a \textit{hazard analysis} is performed before commissioning \cite{STD_ISO10218,STD_ISO12100}.
\begin{figure}
	\centering
	\includegraphics[width=1\columnwidth]{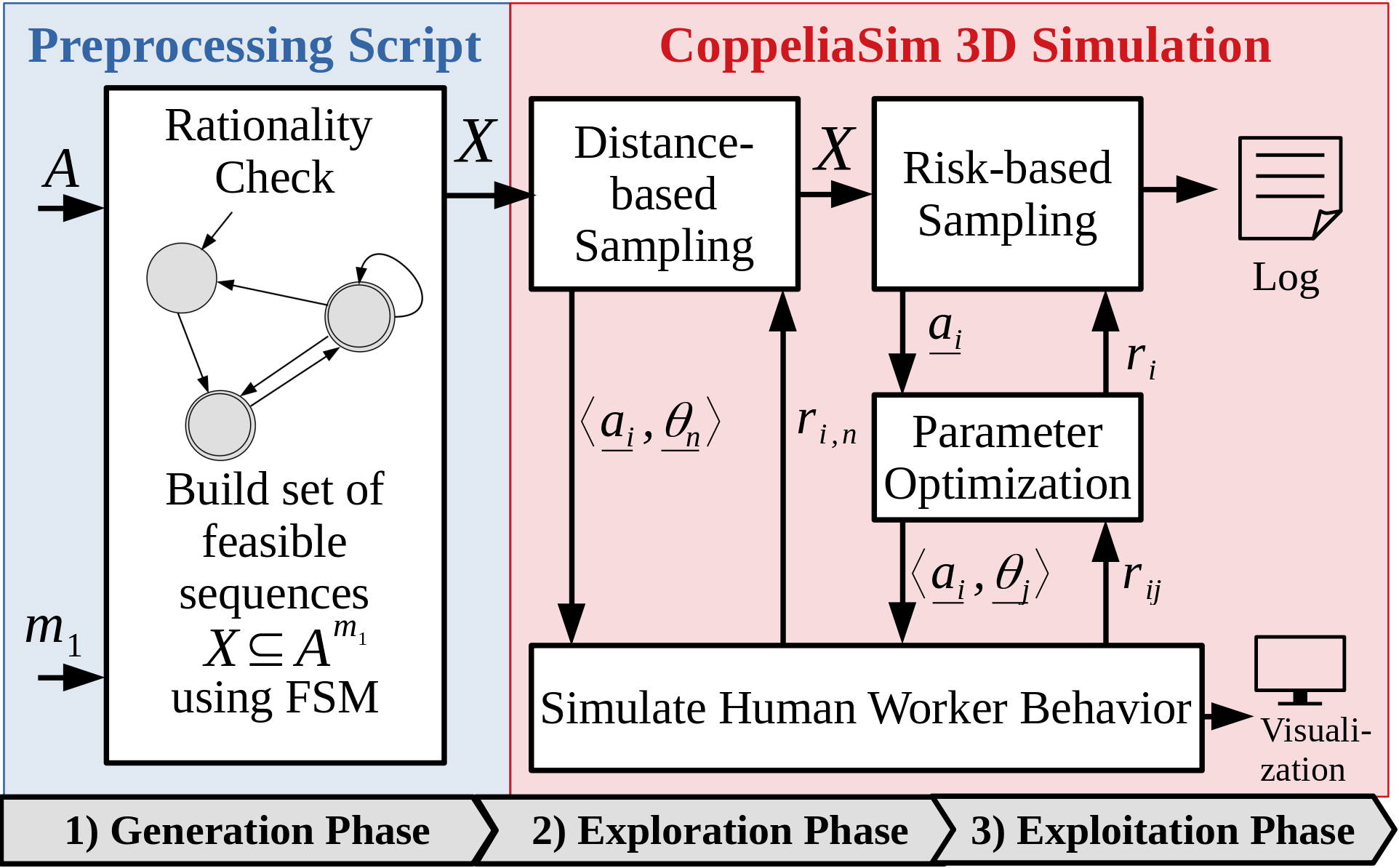}
	\caption{\small Worker behavior is modeled with action sequences $\underline{a}$ and parameters $\underline{\theta}$. Hazardous behaviors are found by: (1) Building a set $X$ of feasible action sequences (Generation), (2) simulating sequences $\underline{a}_i \in X$ with nominal parameters $\underline{\theta}_n$ to get an initial risk estimate $r_{i,n}$ (Exploration), and (3) using optimization to find motion parameters that further increase risk (Exploitation).}
	\label{fig:Structure}
\end{figure}
Current hazard analyses are mostly based on human reasoning, expert knowledge, and experience. In recent years, new approaches including formal verification\cite{RA_Askarpour2016,RA_Vicentini2019} and rule-based expert systems \cite{RA_Awad2017,RA_Wigand2020} have been proposed. While these approaches are viable, they typically operate on rather abstract models. Thus, their effectiveness is limited for systems that are complex or require fine-grained analysis. In such cases, \textit{simulation-based testing} \cite{RA_Araiza2016,RA_Huck2020} is useful, as it can represent certain effects (e.g. collisions) more accurately. 
However, simulation of HRC requires simulating the human worker, which is challenging. Given the variability of human behavior and the possibility of human error, it is insufficient to simulate only the expected worker behavior. Instead, effective simulation-based tests should also cover deviating behaviors to see if they create unsafe states. In complex systems, it may not be immediately obvious what types of behaviors are potentially hazardous. Furthermore, as worker behavior is complex and the number of simulation runs 
is limited by the computational budget, exhaustive simulation of all possible behaviors is infeasible. This leads to the search problem of finding hazardous worker behaviors given a limited number of simulation runs. We address this problem as follows:\newline
\indent \textit{1) Rationality checks}: We restrict the search space by rejecting action sequences that violate workflow-constraints.\newline
\indent \textit{2) Prioritization:} If the remaining sequences are still too many for exhaustive simulation, we prioritize sequences closer to the nominal behavior based on a distance metric.
\newline
\indent \textit{3) Risk-guided search:} We use a risk metric to guide the search towards high-risk behavior, which increases the chances of uncovering hazards.\newline
Note that this paper is not concerned with the development of digital human models. Instead, we assume a human model as given and focus on the aforementioned search problem.\newline
Fig. \ref{fig:Structure} shows the general outline of our approach, which is structured in three phases: In the generation phase, a preprocessing script builds a set of feasible sequences using workflow constraints that are modeled in a finite state machine (FSM). The sequences are fed into the 3D-simulator \textit{CoppeliaSim}\footnote{https://www.coppeliarobotics.com/} to search for the occurence of hazardous states. The search itself is structured in an exploration-phase, where initial risk estimates for the sequences are obtained, and an exploitation phase that attempts to find especially high-risk executions of the sequences (details in Sec. \ref{sec:solutions} and \ref{sec:Implementation}).

\section{Background and Related Work}
\subsection{Robot Safety and Hazard Analysis in Industrial Practice}
\label{sec:IndustrialPractice}
Safety requirements for industrial robot systems are specified by ISO/TS 15066 \cite{STD_ISOTS15066} and \mbox{ISO 10218 \cite{STD_ISO10218}} (the former is intended specifically for HRC). Depending on the robot's capabilities and mode of operation, safety can be achieved through different measures like protective stops, speed and separation monitoring, collision force limitation, or combinations thereof. To implement these measures, collaborative robots (e.g., the \textit{KUKA LBR iiwa} or the \textit{Universal Robots UR10e}) typically provide a set of safety functions such as software-based workspace- and speed-limitations which are programmed and parameterized by users or system integrators according to their specific needs. Additionally, external safety sensors (e.g. light curtains, laser scanners, camera systems) are often used to monitor robot vicinity for approaching humans \cite{HRC_Wang2020}. Choice and configuration of safety measures are highly use-case dependent and influenced by various system-specific factors (e.g., cell layout, reachability of hazardous areas, robot stopping time/distance etc.) \cite{STD_ISO13855}.
To ensure that all relevant hazards are properly addressed, HRC systems are subjected to a hazard analysis\footnote{Note that there are also other terms for this, e.g. "hazard and risk analysis" or "risk assessment". Their meanings differ slightly. For simplicitly we only use the term "hazard analysis".} prior to commissioning \cite{STD_ISO10218}. This is a multi-step procedure to identify risks, estimate and evaluate potential hazards, and determine appropriate safety measures. Standards explicitly require that the hazard analysis does not only consider the nominal workflow, but also erroneous worker behavior \cite{STD_ISO12100}. In current practice, hazard analysis is largely based on human reasoning, expert knowledge, experience, and simple tools (e.g., checklists) \cite{STD_ISO12100,STD_ISO14121}.

\subsection{Novel Hazard Analysis Methods}
\label{sec:NovelMethods}
Current hazard analysis methods are not well-suited for complex (HRC) systems. Thus, a variety of novel methods have been proposed in scientific literature. We shortly introduce the most prominent approaches. For a comprehensive review, we refer to \cite{RA_Huck2021b}.

\subsubsection{Semi-formal Methods}
These methods use semi-formal (often graphic) system representations such as control structure diagrams \cite{RA_Leveson2012}, UML diagrams \cite{RA_Guiochet2013} or task ontologies \cite{RA_Marvel2014} to help the user analyze system behavior and identify potential hazards. Often, they also use a set of guide words to guide the user through the procedure \cite{RA_Leveson2012,STD_IEC61882}. While these methods provide a certain level of support, they have a limited potential to be automated, since core aspects of the hazard analysis are still based on human reasoning.

\subsubsection{Formal Verification}
Formal verification methods automatically check a system model in a formal language for possible violation of safety criteria, allowing for a certain extent of automation in the hazard analysis. In industrial HRC, this approach is mainly pursued by the \textit{SAFER-HRC}-method, which is based on the formal language \textit{TRIO} and the satisfiability checker \textit{ZOT} \cite{RA_Askarpour2016}. The method has also been extended to cover erroneous worker behavior \cite{HUM_Askarpour2017} and can be used in conjunction with a 3D simulator \cite{RA_Askarpour2020} for visualization. Besides industrial HRC, formal safety verification has also been applied to robot systems for personal assistance, medicine, and transport \cite{RA_Webster2014,RA_Choi2021,RA_Proetzsch2007}.
 
\subsubsection{Rule-based Expert Systems}
Rule-based expert systems also provide a certain degree of automation. They use a domain-specific model, such as the \textit{Product Process Resource} model (PPR) \cite{MISC_Ferrer2015}, to describe safety-related workspace characteristics (e.g. layout, which tools are used, and which tasks are performed). Pre-defined rule-sets map these workspace characteristics onto a set of potential hazards. Additional rule-sets can assist the user with adjacent tasks such as risk estimation or decisions about appropriate safety measures \cite{RA_Awad2017,RA_Wigand2020}.

\subsection{Simulation-based Safety Testing}
\label{sec:SimulationBasedTesting}
The methods from Sec. \ref{sec:NovelMethods} are based on relatively abstract system models (e.g., control structure diagrams, formal language descriptions, PPR models) that often require strong modeling simplifications and thus, their applicability is limited when it comes to detailed effects such as human motions or human-robot collisions. These effects can be explored better through \textit{simulation-based testing}. But as simulation models are often complex and safety-critical states rare, creating interesting and critical test scenarios becomes a challenge \cite{FA_Lee2020}. Often, one uses algorithms for search, optimization, or machine-learning to find simulation conditions that expose hazards of failures \cite{FA_Corso2020Survey}. Different heuristics (e.g. risk metrics or coverage criteria) are used to guide the algorithms \cite{FA_Alexander2015,FA_Corso2019RewardAugmentation,FA_Klischat2019}.
So far, simulation-based safety testing has mainly been applied to autonomous vehicles \cite{FA_Babikian2020,FA_Ding2020,FA_Karunakaran2020,FA_Groza2017} and aerospace systems \cite{FA_Lee2015,FA_Alexander2006}.\newline 
In robotics, simulation-based testing has been used mainly for verification and validation on the level of individual system components (e.g., controller- or code-validation) \cite{RA_Araiza2016,FA_Uriagereka2019,FA_Ghosh2018}. Recently, however, the idea of using simulation-based testing as a tool for hazard analysis has also gained some interest. In our earlier work, we proposed the use of Monte Carlo Tree Search (MCTS) to find hazardous states in HRC simulation \cite{RA_Huck2020,RA_Huck2021}. A similar concept was explored in \cite{RA_Andersson2021}, but with Deep Reinforcement Learning instead of MCTS. Other recent publications also consider simulation-based testing as a part of their hazard analysis strategies \cite{RA_Camilli2021,RA_Lesage2021}. Despite this recent interest, the potential of simulation-based testing for hazard analysis in HRC is still relatively unexplored, especially when compared to the wide-spread use that simulation-based testing methods have found in other domains such as autonomous vehicles.

\section{Goal and Problem Definition}
\label{sec:ProblemDefinition}
\begin{figure*}
	\centering
	\includegraphics[width=0.9\textwidth]{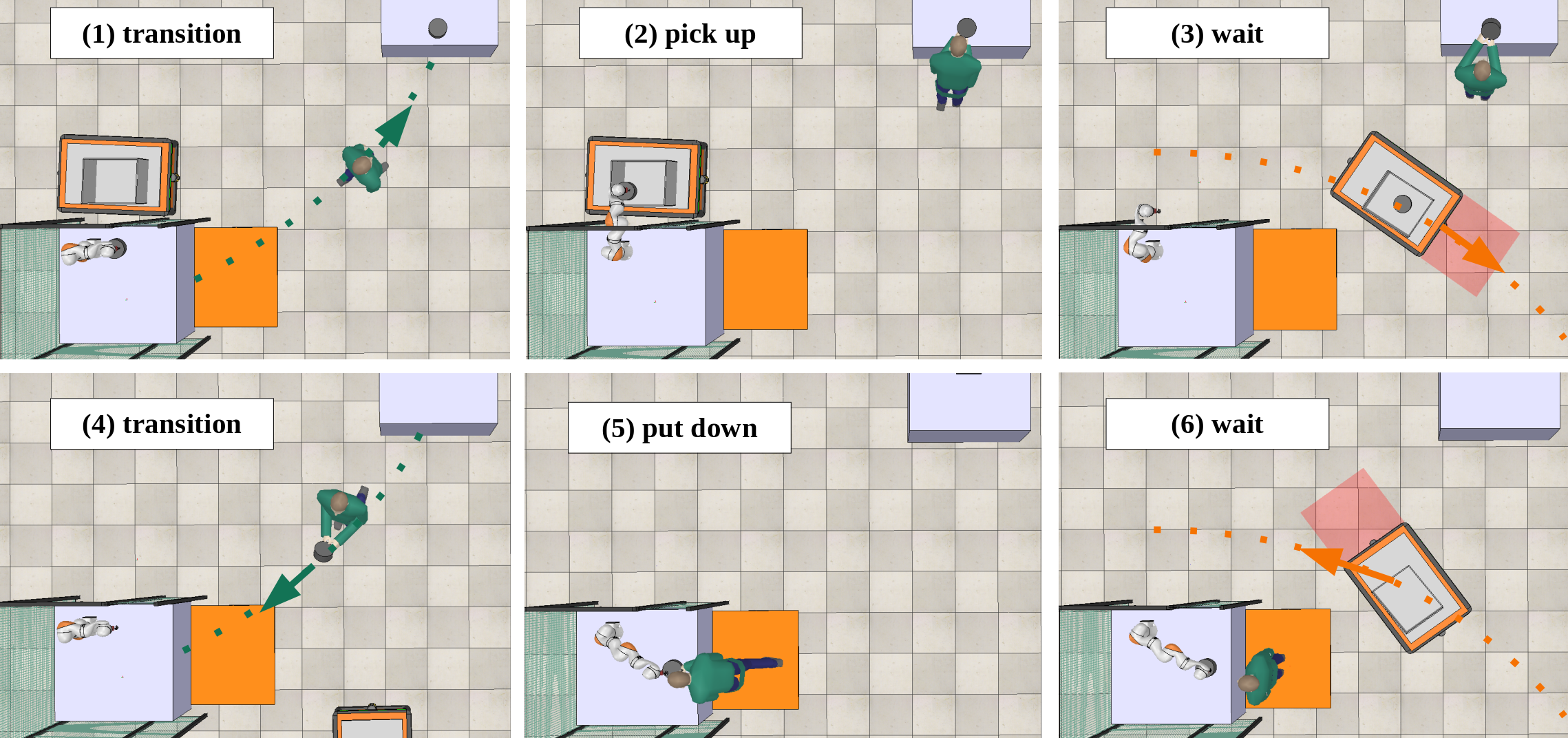}
	\caption{\small A cycle of a simple example workflow: The worker transitions from the robot to the storage station (1), picks up a part (2), waits for the AGV to pass (3), transitions back (4), puts down the part in front (5) and waits for the AGV to return (6).}
	\label{fig:ExampleWorkflow}
\end{figure*}

\subsection{Hazard Analysis as a Search Problem}
This paper focuses on simulation-based safety testing for HRC systems and, more specifically, on the simulation of worker behavior. Our goal is to expose potential hazards by subjecting simulation models of HRC systems to critical worker behaviors that induce unsafe states. We frame this problem as a \textit{search problem} with the goal of finding hazardous behaviors in the search space of possible worker behaviors. We assume that all non-human entities (e.g. robots, sensors, etc.) react deterministically for a given worker behavior\footnote{This assumption may seem very restrictive at first, but is justified since industrial robots are usually pre-programmed and robust to external influences. For significant non-deterministic effects (e.g. varying payloads and stopping distances), a conservative worst-case assumption can be made.}. Assuming a fixed initial state, the behavior of the worker can thus be considered as the only variable on which the resulting simulation states depend. Under this assumption, we can express the search problem as a 4-tuple:
\begin{equation}
	\langle \mathcal{M}, s_0, B, U \rangle
\end{equation}
where $\mathcal{M}$ is the simulation model, $s_0$ the initial simulation state, $B$ the set of possible worker behaviors, and $U$ a (user-defined) set of unsafe states. The goal is now to \textit{find hazardous behaviors} $\tilde{b} \in B$ for which a simulation starting in $s_0$ results in an unsafe state $s_U$:
\begin{equation}
	s_U = \mathcal{M}(s_0,\tilde{b}),\quad s_U \in U
	\label{eq:hazardousBehavior}
\end{equation}
\subsection{Modeling of Worker Behavior}
Human behavior can be modeled on different levels from abstract action sequences to concrete motions \cite{MISC_Bobick1997}. In case of a human worker, unsafe states may result from deviations on a workflow-level (e.g., leaving out worksteps or switching their order), on a motion-level (e.g., walking unexpectedly fast or slow), or from combinations of both. We therefore model worker behavior as a tuple $b=\langle \underline{a},\underline{\theta} \rangle$ consisting of an action sequence $\underline{a}$ and a parameter vector $\underline{\theta}$:
\begin{align}
	\text{with}\quad \underline{a}&=(a_1,a_2,...,a_{m1})\quad a_i\in A\\
	\text{and}\quad \underline{\theta} &= (\theta_1,\theta_2,...,\theta_{m2})\quad \theta_i\in \mathbb{R}
\end{align}
where the action sequence $\underline{a}$ denotes the order of worksteps and the parameter vector $\underline{\theta}$ defines how these worksteps are executed on the motion level (i.e., aspects like walking speed, goals of reaching motions, etc.). Further, $A$ denotes the worker's action space (i.e., the set of all possible worksteps), $m_1$ the length of the action sequence, and $m_2$ the number of parameters. We also assume that there is a known \textit{nominal behavior}, consisting of a nominal action sequence $\underline{a}_n$ and nominal parameters $\underline{\theta}_n$. The nominal behavior represents the behavior that is intended and expected in a normal, non-erroneous action sequence.
With this worker model, our search problem has a mixed discrete-continuous search space of possible behaviors $B$:
\begin{equation}
	B=A^{m_1} \times \mathbb{R}^{m_2}
	\label{eq:SearchSpace}
\end{equation}

\subsection{Example}
\label{sec:Example}
To make the previous definitions more concrete, consider the example in Fig. \ref{fig:ExampleWorkflow}. A worker, a stationary robot, and an automated guided vehicle (AGV) collaborate in a cyclic workflow. The worker's task is to deliver unmachined workpieces to the robot, which processes them and places them on the AGV that carries them away. This means that one cycle is as follows: Initially, the worker stands in front of the robot as a workpiece is processed. After processing, the robot loads the workpiece on the AGV, which takes it away. Meanwhile, the worker transitions to another station (Fig. \ref{fig:ExampleWorkflow}-1) and picks up a new workpiece (Fig. \ref{fig:ExampleWorkflow}-2). The worker waits until the AGV has cleared the area (Fig. \ref{fig:ExampleWorkflow}-3), transitions back to the robot (Fig. \ref{fig:ExampleWorkflow}-4) and puts down the new workpiece in front of the robot (Fig. \ref{fig:ExampleWorkflow}-5). The robot senses the worker's arrival via a sensor mat (orange area in Fig. \ref{fig:ExampleWorkflow}), moves towards the workpiece and starts the processing routine. Meanwhile, the AGV returns to its initial position. Action space and nominal action sequence are:
\begin{align}
	A&=\left\{ t,p,w,d \right\} \label{eq:actionspace}\\
	\underline{a}_n&=(t,p,w,t,d,w) \label{eq:nominalSequence}
\end{align}
with $t,p,w$ and $d$ denoting the actions \textit{(transition)}, \textit{(pick up part)}, \textit{(wait)} and \textit{(put down part)}, respectively. Possible parameters include the worker's walking speed $v$ and the workpiece placement coordinates $x,y$ on the table\footnote{Depending on the desired level of detail, other parameters would be possible as well, e.g. waypoints to describe the worker's walking path. Here we only consider these three parameters for the sake of simplicity.}:
\begin{equation}
	\underline{\theta}=(v,x,y)
	\label{eq:parameters}
\end{equation}
Note that there are $|A|^n=4^6=4,096$ potential action sequences. For each sequence, there is a continuous parameter space in $\mathbb{R}^3$. Even with a relatively coarse parameter discretization into five steps, the search space would still consist of $4096\cdot 5^3= 512,000$ possible behaviors. 

\section{Proposed Solutions}
\label{sec:solutions}
As the example shows, the search space (i.e., the space of possible worker behaviors) can be vast, even in simple models. Exhaustively simulating all possible behaviors is infeasible. Instead, we propose the following three measures:

\subsection{Rationality Checks based on Workflow-Constraints:} \label{sec:exclude} 
\begin{figure}
	\centering
	\includegraphics[width=0.7\columnwidth]{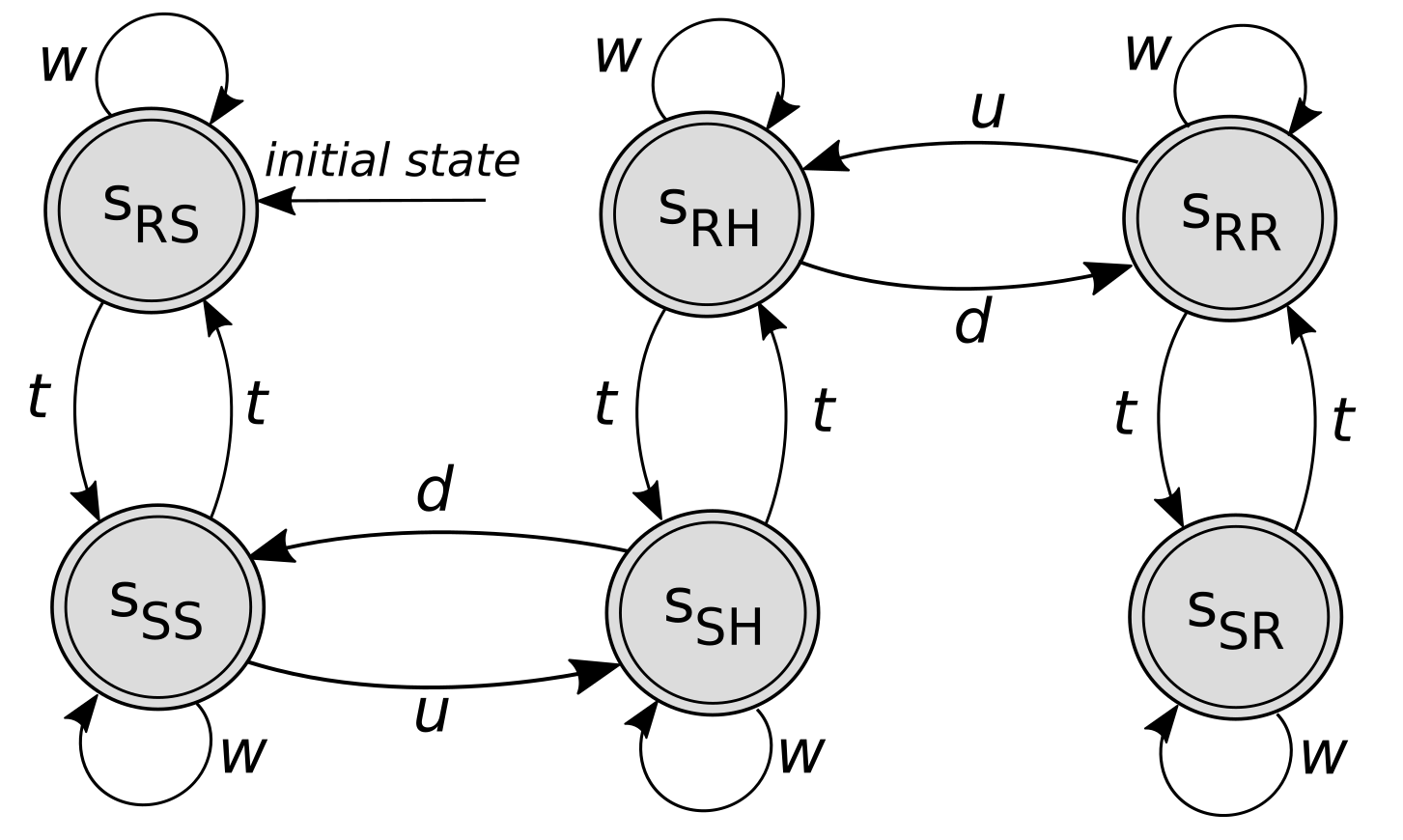}
	\caption{\small The finite state machine determines the feasible action sequences. Note: The first index of the states referes to the worker's position (R: at robot station, S: at storage station) and the second index to the workpiece position (R: at robot station, S: at storage station, H: in worker's hands)}
	\label{fig:FSM}
\end{figure}

It is usually not the case that any action can take any place in a workflow. Instead, there are certain constraints (e.g., a worker cannot put down a workpiece before picking it up). Using such constraints, we can reduce the search space by excluding infeasible sequences a priori (i.e., before running any simulations). We use a Finite State Machine (FSM) to define which actions are possible in which state of the workflow. The set of feasible action sequences $X\subseteq A^{m_1}$ is the set of all action sequences that are accepted by the FSM (here, "accepted" means that it must be possible to run the full action sequence in the FSM without ending up in a state where the required action is impossible).\newline
For an example, consider Fig. \ref{fig:FSM}, wich shows the FSM corresponding to the example workflow from Sec. \mbox{\ref{sec:Example}}. The sequence $(t,u,w,t,d,w)$, for instance, is accepted, whereas the sequence $(t,d,u,w,t,d)$ is infeasible, because the second transition ($d$: put down) is impossible: The worker cannot put down the workpiece in state $S_{SS}$ because it has not been picked up previously.

\subsection{Distance-based prioritization of action sequences:}\label{sec:DistanceMetric} The second measure is to prioritize sequences that are closer to the nominal sequence $\underline{a}_n$. In sequential workflows, erroneous behavior typically results from certain basic errors being superposed to $\underline{a}_n$ \cite{HUM_Hollnagel1993}. Thus, the majority of erroneous sequences are likely to occur in a certain vicinity around $\underline{a}_n$, while highly erratic sequences can be regarded as unlikely. If the computational budget is not sufficient to simulate all feasible sequences, it is therefore reasonable to concentrate on exploring the vicinity of $\underline{a}_n$. To quantify how far a given sequence $\underline{a}$ is from $\underline{a}_n$, we define a distance metric. Based on \cite{HUM_Hollnagel1993}, we assume four basic error types, namely (i) \textit{insertion} of an action into the sequence, (ii) \textit{omission} of an action, (iii) \textit{substitution} of an action by another, and (iv) \textit{reversing} the order of two adjacent actions\footnote{Note that \cite{HUM_Hollnagel1993} lists more than four error types. However, the additional types are special cases of these four.}  (Note: Since we assume a fixed action sequence length $m_1$, sequences subjected to an insertion are truncated to length $m_1$ by dropping the last action, and sequences subjected to an omission are extended to length $m_1$ by inserting an arbitrary action $a\in A$ at the end). We define the distance between $\underline{a}$ and $\underline{a}_{n}$ as the minimum number of basic errors required to convert $\underline{a}_n$ into $\underline{a}$. We call this the \textit{error distance} $d_{e}(\underline{a},\underline{a}_n)$.\\
For an example, consider $\underline{a}_{n}=(t,u,w,t,d,w)$. The following sequence $\underline{a}_1$ results from switching the last two actions in $\underline{a}_{n}$, whereas the action sequence $\underline{a}_2$ results from three action substitutions:
\begin{align}
	\underline{a}_1&=(t,u,w,t,w,d), &\rightarrow\ d_{e}(\underline{a}_1,\underline{a}_{n})=1\\
	\underline{a}_2&=(t,w,w,u,w,w), &\rightarrow\ d_{e}(\underline{a}_2,\underline{a}_{n})=3
\end{align}
Note that the definition of $d_e$ is analogous to the Damerau-Levenshtein distance commonly used to measure the dissimilarity between character strings \cite{MISC_Damerau1964}. Thus, we can use an existing algorithm\footnote{http://www.ccpa.puc-rio.br/software/stringdistance/} to efficiently calculate $d_e$.

\begin{figure*}[h!]
	\centering
	\includegraphics[width=1\textwidth]{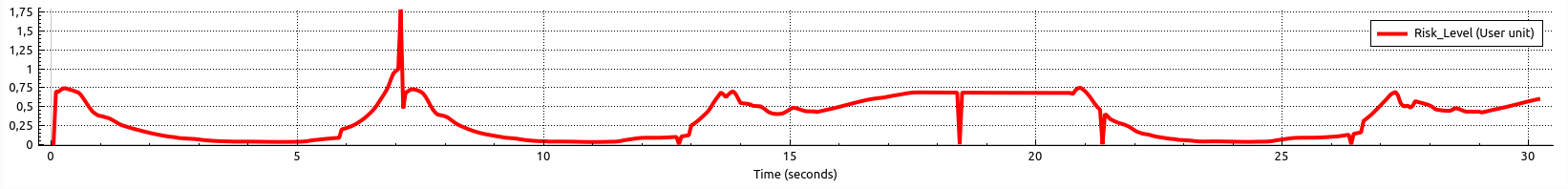}
	\caption{\small Example of the risk level $r$ over simulation time. The high peak corresponds to a collision, the low peaks to protective stops.}
	\label{fig:RiskGraph}
\end{figure*}

\subsection{Risk-guided search:} Thirdly, we use a \textit{risk metric} to guide the search towards high-risk behaviors. 
To account for the mixed discrete-continuous search space (compare Eq. (\ref{eq:SearchSpace})), we use a two-stage search algorithm where the top level samples action sequences guided by the risk metric and the bottom level performs parameter optimization to find high-risk executions of a given action sequence (more in Sec. \ref{sec:Implementation}). Since common risk metrics from safety standards (e.g. \cite{STD_ISO12100,STD_ISO14121}) are typically based on human judgement, they are difficult to calculate automatically. Thus, we define a simplified heuristic metric $r$ to quantify the risk:
\begin{equation}
	r = \begin{cases} 0 & \text{case (a):}\ v_R < v_{crit} \\
		\text{e}^{-d_{HR}} & \text{case (b):}\ v_R \geq  v_{crit};\ d_{HR}>0\\
		\frac{F_{\mathrm{c}}}{F_{\mathrm{max}}}+1 & \text{case (c):}\ v_R \geq  v_{crit};\ d_{HR}=0
	\end{cases}
	\label{eq:RiskMetric}
\end{equation}
Here, $d_{HR}$ denotes the human-robot distance, $v_R$ denotes the robot speed\footnote{for articulated robot arms, $v_R$ is the cartesian speed of the fastest joint; for AGVs, $v_R$ is the robot's translational ground speed},
$v_{crit}$ is a user-defined speed threshold above which the robot is considered potentially dangerous (a typical choice is 250mm/s, see \cite{STD_ISO10218}).
$F_{c}$ denotes the estimated human-robot collision force, and $F_{max}$ is a body-region specific collision force limit \cite{STD_ISOTS15066}.\\
The metric differentiates between three types of situations:
\begin{itemize}
	\item[(a)] The robot is slower than $v_{crit}$ and thus poses a negligible danger to the worker; the risk is zero.
	\item[(b)] The robot is faster than $v_{crit}$, but the distance $d_{HR}$ is greater than zero. Thus, there is no contact (yet), but the robot poses a potential danger to the worker. In this case, the risk is defined on the basis of $d_{HR}$.
	\item[(c)] The robot speed is faster than $v_{crit}$ and the robot is in contact with the worker (i.e. $d_{HR}=0$). In this case, the risk is defined as the ratio of the collision force $F_c$ to the limit $F_{max}$. The addition of +1 ensures that case (c) always returns higher risk than case (b).
\end{itemize}
For a given behavior $b\in B$, the risk is obtained by simulating $b$, evaluating $r$ in each simulation step (see Fig. \ref{fig:RiskGraph}), and returning the highest value that has occurred. It is assumed that exactly one human worker and one or multiple robots are present. For multiple robots, $r$ is evaluated separately with respect to each robot and the maximum value is taken.\newline
Since safety constraints in HRC are typically defined in terms of distance, velocity, and contact forces \cite{STD_ISOTS15066}, the definition of $r$ allows us to express the set of unsafe states $U$ via a risk threshold $r_{th}$:
\begin{equation}
	U=\left\{s\ |\ r>r_{th}\right\}
\end{equation}
For instance, with $r_{th}=1$ all contact situations at critical robot velocity are deemed unsafe, whereas with $r_{th}=2$ only collisions exceeding $F_{max}$ are deemed unsafe.

\section{Search Method}
\label{sec:Implementation}
\subsection{Assumptions, Objective, and General Approach}
This section describes a concrete implementation of this approach. We assume that $A$, $\mathcal{M}$, and $s_0$ are given and that the lengths of action sequences and parameter vectors are $m_1$ and $m_2$, respectively. We also assume that the nominal behavior $\langle \underline{a}_{n}$, $\underline{\theta}_n \rangle$, parameter limits $\underline{\theta}_{min},\underline{\theta}_{max}$, and risk threshold $r_{th}$ are given. The objective of the implemented search method is to find as many hazardous sequences as possible. According to Eq. (\ref{eq:hazardousBehavior}), a sequence $\underline{\tilde{a}}$ is hazardous if parameters $\underline{\tilde{\theta}}$ exist so that the behavior $\tilde{b}=\langle \tilde{\underline{a}},\tilde{\underline{\theta}}\rangle$ results in an unsafe state $s_U\in U$ (i.e., a state where $r_{th}$ is exceeded). 
As seen in Fig. \ref{fig:Structure}, the general procedure is structured in three phases: In the \textit{generation phase}, a preprocessing script builds a set $X$ of feasible action sequences. These are then fed into the 3D Simulator CoppeliaSim. In the \textit{exploration phase}, CoppeliaSim simulates the sequences with their nominal parameters to obtain an initial risk estimate. In the \textit{exploitation phase}, CoppeliaSim deploys a parameter optimizer to find parameters that further increase the risk of the previously explored sequences. We assume that the computational budget is limited in terms of the maximum number of simulation runs $N_{max}$ which are split between exploration and exploitation phase by a split factor $\alpha\in [0,1]$, with $N_{explore}=\lfloor \alpha\cdot N_{max}\rfloor$ simulation runs for exploration and the remaining runs for exploitation.

\subsection{Generation Phase}
The preprocessing script generates a set $X$ of feasible action sequences using the FSM introduced in Sec. \ref{sec:exclude}. The script iterates through the cartesian product $A^{m_1}$ of the action space and checks for each $\underline{a} \in A^{m_1}$ if it is accepted by the FSM. In Algorithm \ref{alg:Pseudocode}, this is represented by ll. \ref{alg:GenerationStart}-\ref{alg:GenerationStop} (Note: The function \textsc{isFeasible} in l. 3 represents the acceptance check).\newline

\subsection{Exploration Phase}
\label{sec:ExplorationPhase}
In the exploration phase (ll. \ref{alg:ExplorationStart}-\ref{alg:ExplorationEnd}), the budget $N_{explore}$ (i.e., the available number of simulation runs) is allocated according to the split factor $\alpha$. If the number of sequences in $X$ is smaller than $N_{explore}$, the remaining budget is allocated to the exploitation phase. Then, the simulator calculates the distance to $a_n$ for the sequences in $X$ and orders the sequences from low to high distance. (Note: For brevity, both the distance calculation and the sorting are represented by a single function in Algorithm 1, l. 14.) 
The sequences in $X$ are simulated with nominal parameters $\underline{\theta}_n$ until the budget of $N_{explore}$ simulations is exhausted. Thereby, an initial risk estimate $r_{i,n}$ is obtained for each simulated sequence. Due to the ordering, sequences closer to $\underline{a}_n$ are prioritized if $N_{explore}$ is insufficient to simulate \textit{all} sequences in $X$. 

\subsection{Exploitation Phase}
When $N_{explore}$ is reached, the simulator transitions into the exploitation phase (ll.  \ref{alg:ExploitationStart}-\ref{alg:ExploitationEnd}) and deploys a parameter optimizer to find parameters that further increase the risk of the explored sequences. Since $N_{exploit}$ is limited, usually not all sequences can be optimized. Instead, sequences are  prioritized according to their initial risk estimates since high-risk sequences have a better chance of exceeding $r_{th}$.
For the prioritization, two alternatives are implemented:\newline
\textit{1) Strict priority:} Sequences are ordered strictly by decreasing risk estimate (Algorithm 1, l. 18).\newline
\textit{2) Probabilistic priority:} Sequences are sampled with probabilities proportional to their risk estimate (this variant is not shown in Algorithm 1):
\begin{equation}
	P(\underline{a}_i) \propto r_{i,n},\quad a_i \in X
\end{equation}
Note that the latter also includes sequences with a low $r_{i,n}$ which would potentially be left out by strict prioritization.\newline
Sequences whose initial risk estimate already exceeds $r_{th}$ are removed from $X$ (these are already identified as hazardous, so further simulations would be futile).\\
For optimization we use a Nelder-Mead Optimizer (NMO). NMO is chosen because it is a \textit{direct search} method that does not require gradient information. This is necessary because the risk value is obtained from simulation, and not from an analytical function, so gradients are not available. Since NMO is well known, we omit a detailed description and refer to \cite{MISC_Nelder1965}. 
The optimization loop is shown in ll. \ref{alg:OptimizationStart}-\ref{alg:OptimizationEnd}. \textsc{NMO.init($m_2$)} randomly initializes the NMO algorithm and returns an initial parameter vector $\underline{\theta}$. Then, the first sequence is simulated with these parameters, and the resulting risk is given to \textsc{NMO.iterate($r$)}, which performs one step of the NMO algorithm and returns a new $\underline{\theta}$. This is repeated iteratively until $r_{th}$ is exceeded or a maximum number of steps $N_{NMO}$ is reached, in which case the algorithm moves on to the next sequence. Penalty functions ensure that parameters remain in $[\theta_{min},\theta_{max}]$. (Note: For brevity, this aspect is not shown in Algorithm \ref{alg:Pseudocode}.) 
The exploitation phase terminates when the budget $N_{explore}$ is exhausted.
\begin{algorithm}
	\begin{algorithmic}[1]
		\State $X \leftarrow \langle\  \rangle$, $i\leftarrow1$   \label{alg:GenerationStart}
		\For{\textbf{each} $\underline{a}\in A^{m_1}$} \color{gray} \# select feasible sequences \color{black}
		\If {\textsc{isFeasible}($\underline{a}$)}
			\State $X_i.\text{actions}\leftarrow \underline{a}$
			\State $i\leftarrow i+1$
		\EndIf	
		\EndFor \label{alg:GenerationStop}
		\If{$|X|\geq\lfloor \alpha\cdot N_{max}\rfloor$}  \color{gray} \# alloc. exploration budget \color{black}\label{alg:ExplorationStart}
			\State $N_{explore}\leftarrow\lfloor \alpha\cdot N_{max}\rfloor$
		\Else
			\State $N_{explore}\leftarrow |X|$
		\EndIf
		\State $N_{exploit}\leftarrow N_{max}-N_{explore}$ \color{gray} \#remaining budget \color{black}
		\State $X\leftarrow \textsc{sortByIncreasingDistance}(X)$
		\For{$i=1,\lfloor \alpha\cdot N_{max}\rfloor$}\label{alg:sortByIncreasingDistance} \color{gray} \# exploration \color{black}
			\State $X_i.\text{risk} \leftarrow \textsc{simulate}(X_i.\text{actions},\underline{\theta}_n)$
		\EndFor \label{alg:ExplorationEnd}
		\State $X\leftarrow \textsc{sortByDecreasingRisk}(X)$ \label{alg:ExploitationStart}
		\State $X\leftarrow X \setminus \{X_i\ |\ X_i.\text{risk}>r_{threshold}\}$\color{gray} \# remove sequences that already exceed $r_{th}$ \color{black}
		\State $i\leftarrow1,j\leftarrow1$ \color{gray} \# i: sim runs, j: sequences in $X$\color{black}
		\While{$i<N_{exploit}$}
			\State $k\leftarrow1$ \color{gray} \# optimization step counter \color{black}
			\State $\underline{\theta}\leftarrow \textsc{NMO}.\text{init}()$
			\Do \label{alg:OptimizationStart}
				\State $r\leftarrow \textsc{simulate}(X_j.\text{actions},\underline{\theta})$
				\State $\underline{\theta}\leftarrow \textsc{NMO}.\text{iterate}(r)$
				\State $k\leftarrow k+1$
			\doWhile $r < r_{threshold}\ \text{\textbf{and}}\ k<N_{NMO}$ \label{alg:OptimizationEnd}
			\State $j\leftarrow j+1$ \color{gray} \# go to next sequence \color{black}
			\State $i\leftarrow i+k$ \color{gray} \# k sim runs conducted in inner loop \color{black}
		\EndWhile\label{alg:ExploitationEnd}
	\end{algorithmic}
	\caption{Search Method Pseudocode}
	\label{alg:Pseudocode}
\end{algorithm}

\section{Application Example}
\label{sec:Test}
\subsection{Scenario}
\begin{figure}
	\centering
	\includegraphics[width=0.9\columnwidth]{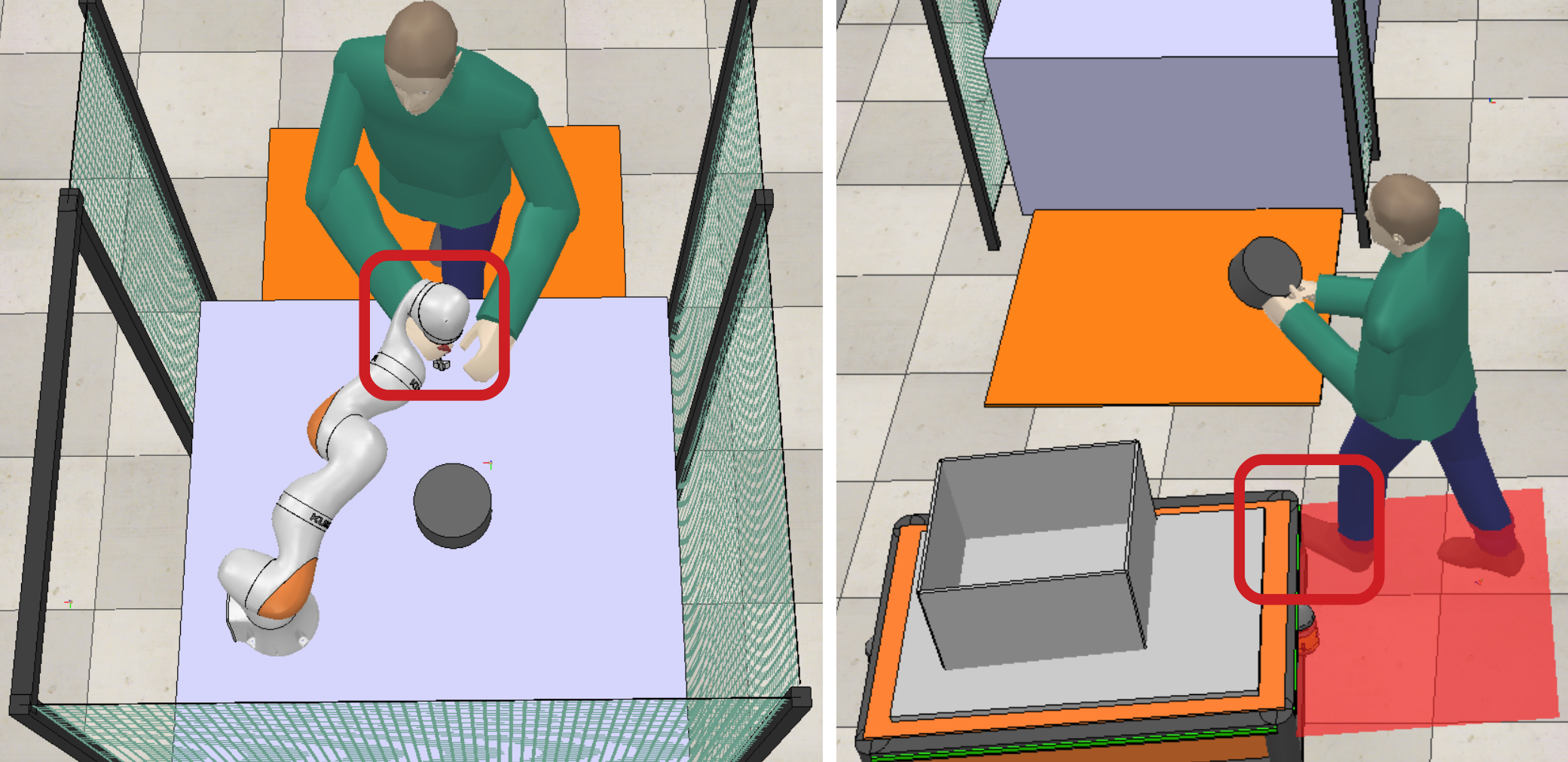}
	\caption{\small Left: The worker's hand collides with the robot after placing the workpiece on the table. Right: Worker and AGV collide.}
	\label{fig:CollisionExamples}
\end{figure}

We present an example using the workflow from \mbox{Sec. \ref{sec:Example}} (a second example with a slightly more complex workflow is available on GitHub\footnote{https://github.com/Huck-KIT/Robot-Hazard-Analysis-Simulation}, but is not discussed here for reasons of brevity). Action space $A$ and parameters $\underline{\theta}$ are chosen according to Eq. (\ref{eq:actionspace}) and Eq. (\ref{eq:parameters}), respectively. The nominal sequence $\underline{a}_{n}$ is chosen according to Eq. (\ref{eq:nominalSequence}). The workflow constraints are given by the FSM shown in Fig. \ref{fig:FSM} (see \mbox{Sec. \ref{sec:exclude}} for details).
Our safety criterion is that there must be no collisions where the contact force exceeds $70\%$ of the collision force limit. Thus, we set the risk threshold to $r_{th}=1.7$ (compare Eq. (\ref{eq:RiskMetric})). The set of unsafe states is:
\begin{equation}
	U=\{s\ |\ r>1.7\}
\end{equation}
For testing purposes, deliberate safety flaws are designed into the system, so that both robot and AGV can cause an unsafe state if the worker behaves in certain ways that deviate from the nominal behavior (see Fig. \ref{fig:CollisionExamples}).
The generation phase yields 266 feasible action sequences from the initially 4096 potential combinations.

\subsection{Test Runs and Results}
\label{sec:TestRuns}
\begin{figure}
	\centering
	\includegraphics[width=0.9\columnwidth]{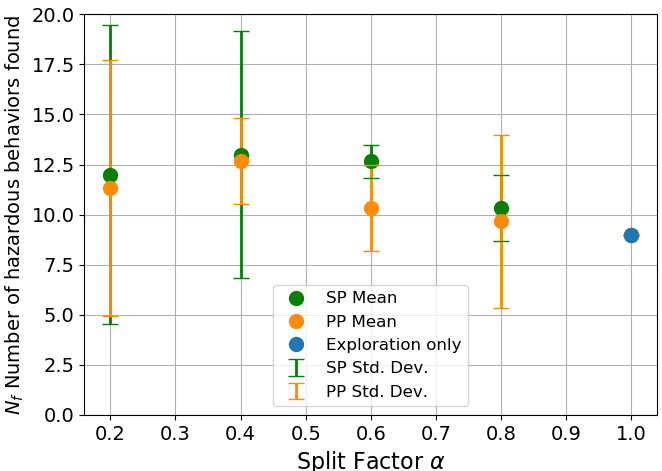}
	\caption{\small Split factor sweep. Search configuration: SP, $N_{max}=266$, $N_{NMO}=8$. For each $\alpha$, three test runs were conducted, except for $\alpha=1$, where the search is fully deterministic (exploration only).}
	\label{fig:ParameterSweep}
\end{figure}

The 266 remaining feasible sequences are searched for hazardous behavior in several different test runs.\newline
\textit{1) Comparison of search configurations:} Test runs with a constant split factor were performed to compare different search configurations: Search with strict prioritization (SP), search with probabilistic prioritization (PP), and, as a baseline, a random search (R) where sequences and parameters are sampled from uniform distributions over $X$ and $[\underline{\theta}_{min},\underline{\theta}_{max}]$ without any form of distance- or risk-prioritization. For each approach, ten test runs were conducted with $N_{max}=300$ and \mbox{$N_{max}=500$}, respectively. The results are reported in Table \ref{tab:results}. For \mbox{$N_{max}=500$}, SP outperforms R in terms of the number of identified hazards, while the error distances of the identified hazards are similar. For $N_{max}=300$, PP slightly underperforms R, but the identified hazards are closer to the nominal behavior than those found by R. Interestingly, PP underperforms both R and SP as it tends to waste simulation time with low-risk sequences that have little chance of exposing any hazard.\newline
\textit{2) Split factor sweep:} Additionally, tests with a fixed search configuration, but different split factors were conducted. For these tests, $N_{max}=266$ was chosen (the same as the number of feasible sequences), so that an exploration-only search corresponds with $\alpha=1$. Results are shown in \mbox{Fig. \ref{fig:ParameterSweep}}. The best results were obtained with a medium split factor ($\alpha=0.4$ and $\alpha=0.6$), which is expected as a small 
$\alpha$ leads to insufficient exploration of the search space, whereas a large $\alpha$ leaves too few simulation runs for optimization. Yet, one should keep in mind that the influence of $\alpha$ is problem-dependent and might look very different in another scenario.

\begin{table}
	\centering 
	\begin{tabular}{p{0.2cm}  p{0.2cm} p{0.2cm} p{0.2cm} p{0.2cm} p{0.2cm} p{0.2cm} | p{0.2cm} p{0.2cm} p{0.2cm} p{0.2cm} p{0.2cm} p{0.2cm}} 
		\toprule 
		& \multicolumn{6}{c}{{Results for $N_{max}=500$}} & \multicolumn{6}{c}{{Results for $N_{max}=300$}}\\
		\midrule
		& \multicolumn{2}{c}{R} & \multicolumn{2}{c}{SP}& \multicolumn{2}{c}{PP}& \multicolumn{2}{c}{R} & \multicolumn{2}{c}{SP} & \multicolumn{2}{c}{PP}\\
		\textbf{\#} & $N_{f}$ & $d_{e}$ & $N_{f}$ & $d_{e}$& $N_{f}$ & $d_{e}$& $N_{f}$ & $d_{e}$& $N_{f}$ & $d_{e}$& $N_{f}$ & $d_{e}$\\ 
		\midrule 
		1 & 27 & 3.1 & 36 & 3.2 & 24 & 3.1 & 20 & 3.2 & 19 & 2.6 & 12 & 2.9 \\ 
		2 & 27 & 3.2 & 38 & 3.2 & 21 & 3.1 & 17 & 3.2 & 19 & 2.8 & 12 & 2.8\\ 
		3 & 24 & 2.5 & 37 & 3.3	& 22 & 2.6 & 9  & 3.2 & 15 & 2.9 & 12 & 2.6\\ 
		4 & 27 & 3.1 & 32 & 3.2 & 26 & 3.3 & 20 & 2.6 & 19 & 2.8 & 10 & 2.8\\ 
		5 & 24 & 2.8 & 37 & 3.2 & 25 & 2.8 & 24 & 2.6 & 15 & 2.4 & 13 & 2.8\\ 
		6 & 25 & 3.3 & 34 & 3.2 & 27 & 2.8 & 16 & 3.1 & 17 & 3.0 & 12 & 2.7\\ 
		7 & 29 & 3.0 & 32 & 3.2 & 23 & 2.7 & 12 & 3.4 & 18 & 2.2 & 10 & 2.8\\ 
		8 & 28 & 3.0 & 33 & 3.2 & 23 & 3.0 & 15 & 2.9 & 14 & 2.5 & 16 & 2.6\\ 
		9 & 27 & 3.1 & 31 & 3.2 & 23 & 3.2 & 18 & 2.8 & 16 & 2.6 & 10 & 2.8\\ 
		10& 29 & 3.4 & 30 & 3.2 & 23 & 2.9 & 21 & 2.8 & 17 & 2.5 & 12 & 2.5\\ 
		\midrule
		\textbf{Avg.} & \textbf{26.7} & \textbf{3.1} & \textbf{34} & \textbf{3.2}& \textbf{23.7}& \textbf{3.0}& \textbf{17.2} & \textbf{3.0} & \textbf{16.9} & \textbf{2.6} &\textbf{11.9} &\textbf{2.7}\\
		\bottomrule
	\end{tabular}
	\smallskip 
	\caption{\small Comparison of search configurations. R: Random Search, SP: Search with strict prioritization, PP: Search with probabilistic prioritization. $N_f$: Number of hazardous behaviors found, $d_e$: Avg. error distance of hazardous behaviors found. In all cases, $\alpha=0.6$ and $N_{NMO}=8$ was chosen.} 
	\label{tab:results}
\end{table}

\section{Discussion and Future Work}
The presented method enables users to consider effects of deviating worker behavior in their hazard analyses. The proposed simulation-based approach has several benefits: Simulation models are often already available in the development phase, so renewed system modeling is largely avoided (except for modeling the FSM to describe workflow constraints). Also, compared to PPR models or formal language descriptions, simulation captures physical effects (e.g. movements or collisions) more accurately. Finally, the simulation is treated similarly to a black-box model: Apart from action space, parameter ranges, and workflow constraints, no explicit knowledge about the systems' internal workings is required (e.g., what safety measures are used or how they are configured). This information is contained \textit{implicitly} in the simulation, but there is no need to formalize it \textit{explicitly} (as it would be the case e.g. in rule-based methods). This improves scalability to complex systems.\\
The main challenge is the large search space of possible worker behaviors. Our test example indicates that the proposed search techniques can effectively cope with this, although more extensive tests in different scenarios are needed to confirm the promising initial results. Note that our approach is limited in the sense that it can only \textit{falsify} (i.e, find cases of safety constraints being violated) but cannot provide a safety guarantee. As with all simulation-based techniques, success depends on the model quality. With faulty or too simplistic models, the simulation might find hazards that do not exist in reality, or, more concerning, real-world hazards might not manifest themselves in simulation. For instance, our example uses a relatively simplistic worker model where certain potentially hazardous behaviors (e.g. deviations in the walking path) are not captured.
Future work will therefore focus on the integration of more sophisticated digital human models and on the use of advanced search algorithms. It would also be interesting to test performance against an exhaustive search rather than a random search as a baseline. In the present paper this was not yet done due to the extensive computation time required.\\
Finally, we emphasize that the proposed method is an \textit{addition} to existing ones and not a replacement. In the long term, simulation-based testing could be used in conjunction with formal verification to provide more detailed analysis of cases where formal verification indicates potential hazards.

\section*{Acknowledgment}
We thank Tamim Asfour for his support and constructive discussions.
\bibliographystyle{IEEEtran}
\bibliography{RA-L2021_Preprint}
\end{document}